# CartoMark: a benchmark dataset for map pattern recognition and map content retrieval with machine intelligence

## Authors
Xiran Zhou[1], Yi Wen[1], Honghao Li[1], Kaiyuan Li[1], Zhenfeng Shao[2], Zhigang Yan[1], Xiao Xie[3]

**Affiliations**
1. School of Environment and Spatial Informatics, China University of Mining and Technology, Xuzhou, China
2. State Key Laboratory of Surveying, Mapping and Remote Sensing, Wuhan University, Wuhan, China
3. Applied Ecological Institute, Chinese Academy of Sciences, Shenyang, China
corresponding author(s): Xiran Zhou (xrzhou@cumt.edu.cn)

## Abstract
Maps are fundamental medium to visualize and represent the real word in a simple and philosophical way. The emergence of the 3rd wave information has made a proportion of maps are available to be generated ubiquitously, which would significantly enrich the dimensions and perspectives to understand the characteristics of the real world. However, a majority of map dataset have never been discovered, acquired and effectively used, and the map data used in many applications might not be completely fitted for the authentic demands of these applications. This challenge is emerged due to the lack of numerous well-labelled benchmark datasets for implementing the deep learning approaches into identifying complicated map content. Thus, we develop a large-scale benchmark dataset that includes well-labelled dataset for map text annotation recognition, map scene classification, map super-resolution reconstruction, and map style transferring. Furthermore, these well-labelled datasets would facilitate the state-of-the-art machine intelligence technologies to conduct map feature detection, map pattern recognition and map content retrieval. We hope our efforts would be useful for AI-enhanced cartographical applications.

## Background & Summary

According to the definition of International Cartographic Association (ICA)[1], a map is defined as "a symbolized representation of geographic reality, representing selected features or characteristics, resulting from the creative effort of its author's execution of choices, and is designed for use when spatial relationships are of primary relevance." Thus, maps are fundamental medium to visualize and represent the real word in a simple and philosophical way. It simplifies the complicated world and connect people to the reality through an inspired imagination[2]. As shown in Figure 1, the map is a visual representation of the reality, or geographical landscape.

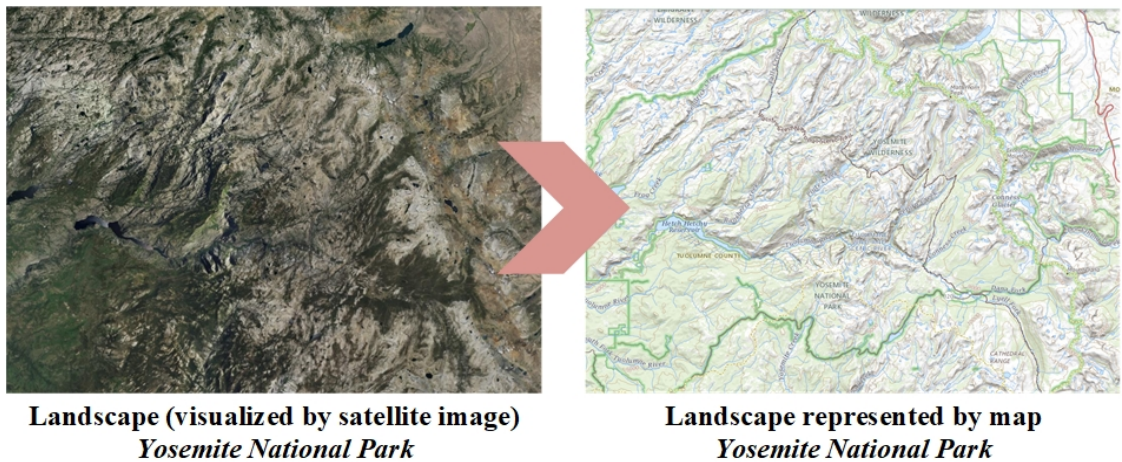

Figure 1 The visual representation of map on real landscape, in case of Yosemite National Park.

Up to now, a variety of maps have been emerged throughout human history and have been identified as one of the three common human being' languages, alongside music and drawing. Researches regarding cartography and socio-economic development have reported that maps hold the ability to represent the natural and social worlds by a scientific aspect (e.g. portraying scientific principles), a cultural aspect (recording the art and cultures), a social aspect (depicting public opinions), a historical aspect (visualizing historical maps status.), a legal aspect (displaying administrative boundaries), and a military aspect (illustrating the historical wars)[3-5]. Thus, the-state-of-the-art machine learning and deep learning approaches have been employed to effectively process and analyze standard maps being generated by professional mapping principles for extracting and mining the geospatial information on a place[6-7].

The emergence of the 3rd wave information has revolutionized the paradigm of cartography and map-related applications, as a majority of maps are generated ubiquitously. This allows people to create, edit, publish and share their map products in respect to a place based on personal viewpoints and subjective understandings. In addition, these massive maps generated by different countries, institutes and people are available for public uses through various platforms including earth observation system, information cyberinfrastructures, and big data techniques[2,8]. The geospatial information stored in these the maps accessed from massive sources are exponentially larger than those stored in professionally-generated maps. Thus, recognizing and retrieving the content of the maps accessed from massive sources, or big map data, would be significance for enriching the dimensions and perspectives to understand the characteristics of the real world, and enhance the capability of geospatial analysis.

However, since a significant proportion of maps have been created by diverse mapping ways that follow non-cartographic rules, their characters, symbols, scenes, and other map features might considerably vary in styles, arrangement and background[2,8,9]. These differences pose a considerable challenge for the-state-of-the-art deep learning approaches to learn salient features in terms of map content. Figure 1 illustrates the results represented by the maps accessed from massive sources: the figure of this sport field is visualized differently by a variety of maps. Thus, the researches of cartography and location-based services have found that a majority of map dataset have never been discovered, acquired and effectively used, leading to the waste of plentiful information contained within these data resources[10,11]. Meanwhile, the map data used in many applications might not be completely fitted for the authentic demands of these applications[8,12].

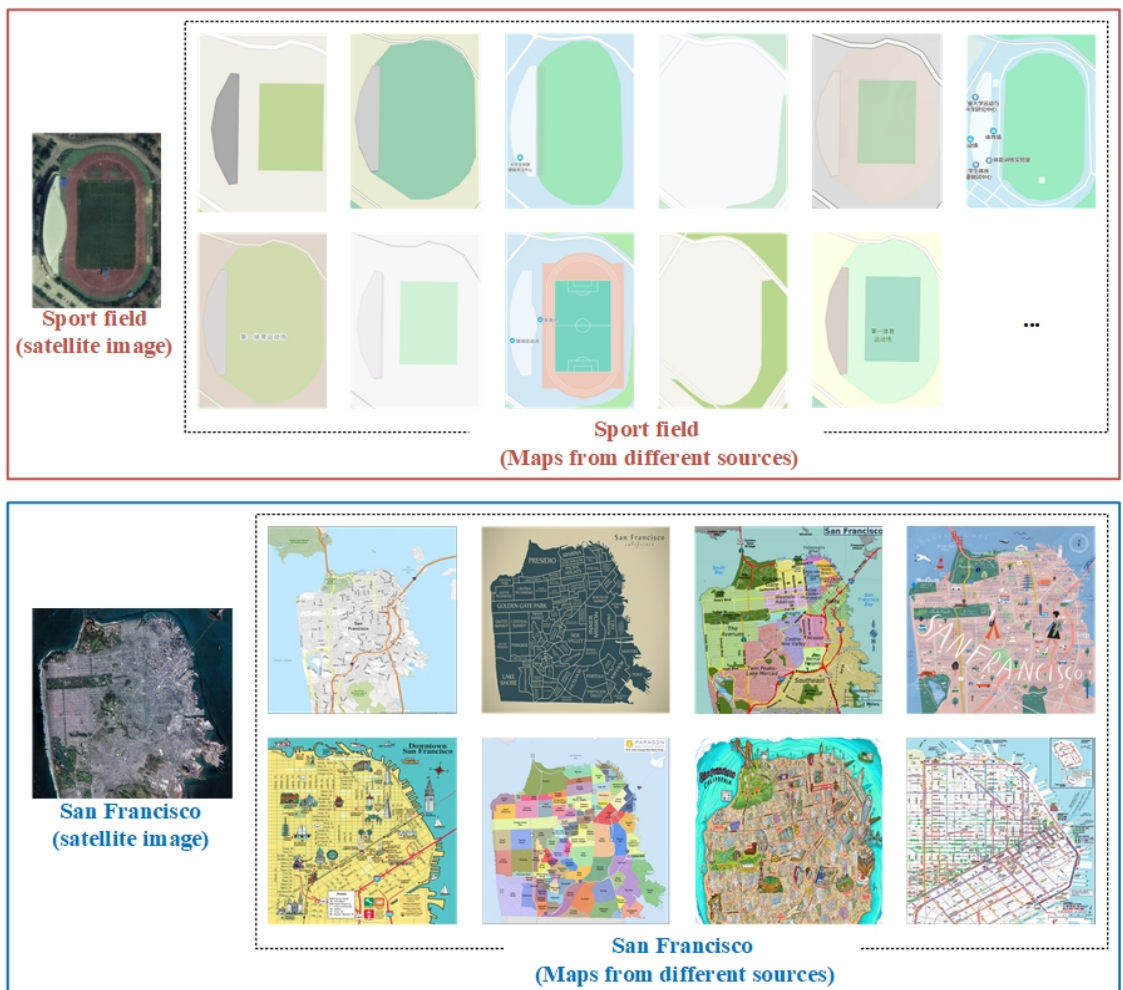

Figure 2 Comparison of maps from different sources: in case of San Francisco and sport field.

Classical non-content-based approaches such as metadata-based map analysis[13] and file name-based map analysis[14] might not be capable of recognizing the content in the maps accessed from massive sources. The metadata and title of the maps accessed from massive sources are generated by subjective cognition, meaning that the same map may be assigned different metadata and names. This limitation has been reported in the implementation of metadata-based and file name-based big map data recognition and retrieval[15]. In the meantime, expert-based intelligent systems have been developed for map content analysis[16]. However, these labor-driven machine-person interactions might be insufficient for the maps accessed from massive sources due to the divergence of various map content. making them impossible to be fully defined[17]. Other approaches including image morphology[18,19] and object-based image analysis[20-22] always requires a manually-defined optimized threshold for precisely extracting map object, which are also a labor-intensive works.

Deep learning techniques has attracted a significant attention from the communities of cartography and geospatial information science because of its powerful capability of feature learning. However, deep learning approaches always rely on numerous well-labelled datasets. The lack of a benchmark dataset makes it impossible to implement the deep learning approaches into identifying complicated map content. Considering maps' potential to enhance a variety of applications in terms of cultural narratives, navigations, location-based services, etc., the gap between sufficient map data and unproductive map utilization pose an urgent demand: developing a large-scale benchmark dataset that facilitate the state-of-the-art machine intelligence technologies to accurately detect map features, recognize map patterns and conduct map content retrieval.

This manuscript reports the details regarding a large-scale benchmark dataset called *CartoMark*. We develop this benchmark dataset to facilitate people utilizing the cutting-edge machine intelligence techniques and testifying their proposed machine intelligence approaches for map pattern recognition and map retrieval.

## Methods

### Data Sources

As mentioned above, maps from ubiquitous sources are always generated by different ways. This means that the data sources are ubiquitous, and mainly include three types of source: Internet, data cyberinfrastructure/repositories, and social media.

The first data source is the Internet, the data of which are available to be accessed by commonly-used search engines including Google image search, Bing image search, and Baidu image search. The second data source is the data cyberinfrastructure/repositories including USGS Earth Explorer, USGS Historical Topographic Map Explorer, and Tianditu Map. Besides the previous two sources, a small proportion of maps are accessed from the third data source—social media including Facebook, Twitter, Instagram and TikTok.

### Architecture of CartoMark

Figure 3 illustrates the architecture of CartoMark, which supports three cartographical pattern recognition tasks: map text annotation recognition, map super-resolution reconstruction, map scene classification, and map style transferring. The dataset for all four tasks could be conducted by the-state-of-the-art machine intelligence techniques. Map text annotation recognition aims to detect and recognizing the text characters in a map. Its repository includes three data groups, each providing map files and the corresponding label files. The map files in different groups are identical in content but in two formats: JPG and PNG. Map super-resolution reconstruction focuses on generating the higher resolution maps from the original map in low resolution. Its repository includes two groups of map files. With the map files in different groups being identical in content but different in format: JPG and PNG. Map scene classification concentrates on classifying maps into various categories based on their content. Its repository includes two data groups, each providing map files and a label file. The map files in different groups are identical in content but different in format: JPG and PNG. Map style transferring provides samples to transfer the original maps into different styles without changing the initial content. Its repository includes two groups of map files, which are identical in content but different in format: JPG and PNG.

Specifically, we provide three two of map formats, considering that the state-of-the-art machine intelligence techniques require one of these three data formats. This can save users' time and avoid the workload of format conversion. Moreover, the results of map text annotation recognition, map super-resolution reconstruction, map scene classification, and map style transferring can further support map retrieval and map semantic mining.

Moreover, we provide three python-encoded programs (or tools) to process the map samples for other possible applications. *Image format conversion* is a tool to convert the available map format into other main formats. *Draw text line* is a tool for visualizing the bounding box that covers a map text. *Text annotation format conversion* is a tool to transform the data structure of .txt files into other data structures required by various deep larnin techniques regarding OCR.

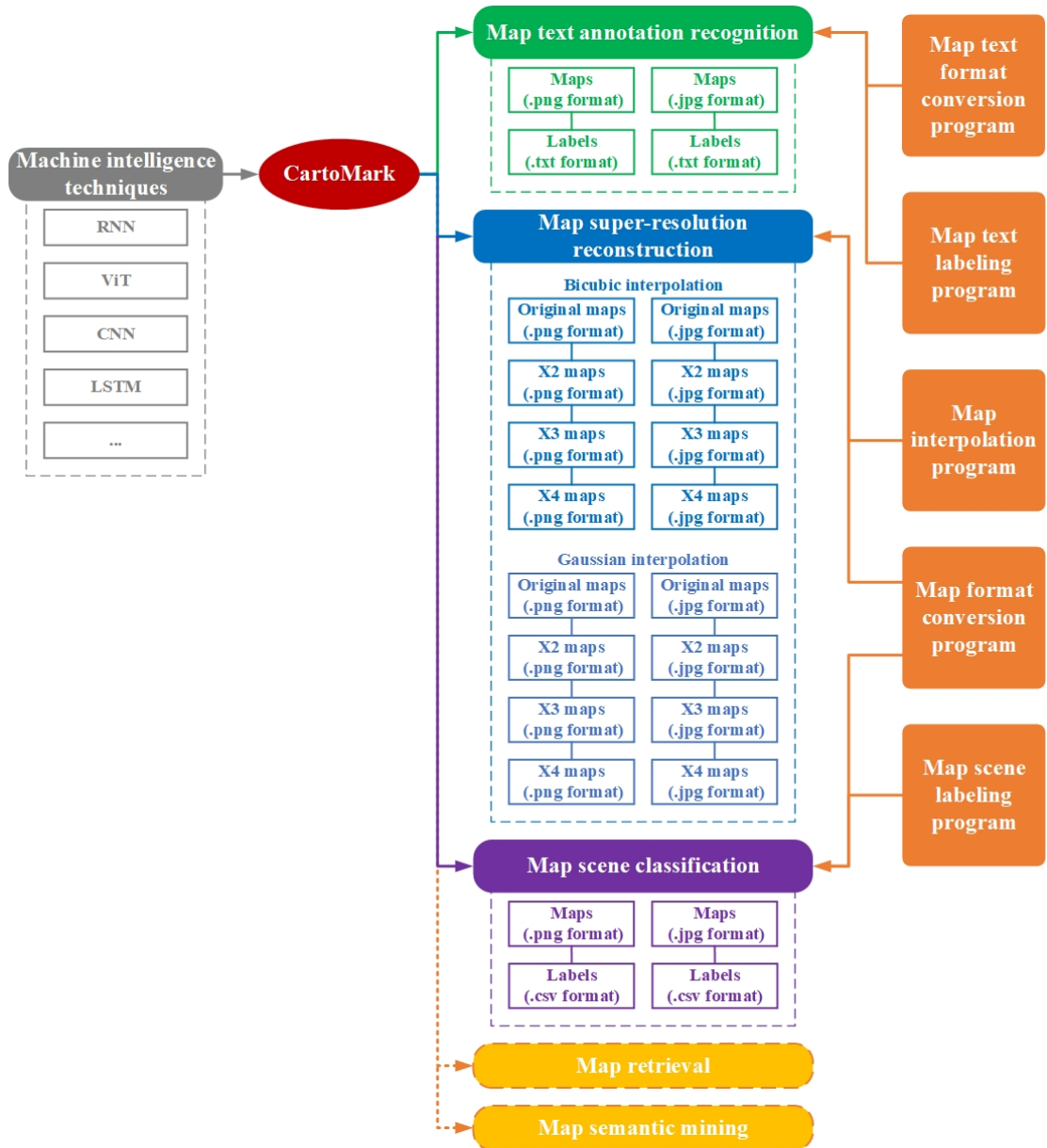

Figure 3. Architecture of CartoMark.

## Taxonomy and folksonomy integrated map retrieval

**(1) Map-dataset harvesting**

As mentioned above, maps from ubiquitous sources are always generated by different ways. This means that the strategy of map-dataset harvesting should employ a variety of ways instead of a single one. Maps generated by professional cartographical rules can be assessed using standardized cartographical taxonomies. However, for the maps accessed from massive sources, a map that holds the similar content might be assigned different names (tags), which belongs to folksonomy in the information retrieval. Thus, we propose the keywords used for retrieving the original map dataset by unifying the taxonomies of professionally-generated maps and the folksonomies of the maps accessed from massive sources. The taxonomy and folksonomy integrated keywords are listed in Table 1.

Table 1. List of the taxonomy and folksonomy integrated keywords

| Keyword categories | Keyword individuals |
|---|---|
| City | Beijing, Shanghai, Guangzhou, Xi'an, Shenzhen, Hangzhou, Nanjing, New York, Washington DC, Los Angeles, San Francisco, Seattle, Chicago, Philophobia, London, Manchester, Paris, Madrid, Barcelona, Berlin, Frankfurt, Munich, Hamburg, Milan, Turin, Roma, Amsterdam, Moscow, Istanbul, Vienna, Prague, Stockholm |
| Continent | North American, African, Asian, Europe, South American |
| Space | three-dimensional, 3D, virtual, bird-eye |
| Nature | hydrology, water net, topographical, terrain, digital elevation model, TIN, digital surface model |
| Social-economy | indoor, shopping mall, airport, museum, cinema, travel, tourism, urban, traffic, street, land classification |
| Others | comic, drawing, |

Based on these keywords, we employed "keyword individual + map" to retrieve and collect map files by commonly-used search engines, data cyberinfrastructure/repositories, and social media crawlers. For example, we used "San Francisco map", "indoor map", "airport map", etc. to retrieve and collect all relevant maps. The commonly-used search engines include Google Image Search, Amazon Image Search, Yahoo Image Search, Bing Image Search, and Baidu Image Search. The data cyberinfrastructure/repositories include USGS Earth Explorer, USGS Historical Topographic Map Explorer, and Tianditu Map. For web crawler, we employed the open-source crawlers to retrieve the map files that holds "keyword individual" and "map" in their file names from the Internet and social media. However, since a majority of social media were not available for public accessing, the amount of maps retrieved from crawlers was limited. Moreover, professional maps accessed from data cyberinfrastructure/repositories always holds similar map styles and configurations. Thus, a majority of the original map dataset were obtained from the web search engines. There are around 20000 maps are collected from different sources, and stored as the original map dataset.

**(2) Map-dataset cleaning and map filtering**

Since a majority of maps being accessed from ubiquitous sources are generated by unprofessional ways, the quality of these maps might be jagged and rough. We propose an integrated strategy to filter and clean the original maps being unqualified as the samples in the benchmark dataset. The strategy includes five steps: image size filtering, image format unifying, image color and texture filtering, image noise removing and image content checking. Figure 4 shows the workflow of this strategy.

1. Map resizing. 1) We set the minimal image size as 256*256, and all maps that smaller than this minimal size would be remove. This is based on the requirement of input data dimension of the state-of-the-art deep learning approaches. 2) For a map that holds the size from 256*256 to 1024*1024, we would crop the central part of this map as the original map. 3) For a map that is larger than 1024*1024, we would randomly partition this original map into four sub-parts, which were used as the original maps. The cropping and partitioning will be conducted according to the following rules.

Assuming the dimension of an original map dataset as $(x, y)$, for the dimension proposal $(d, d = \max(x, y))$,

If $d \in (512, 1024)$, we cropped the central part of this map as the original map. Thus, the area of the cropped map is $(\frac{1}{4}x:\frac{3}{4}x, \ \frac{1}{4}y:\frac{3}{4}y)$.

If $d \in (1024, +\infty)$, we partitioned the original map into two parts, the area of which are: $(0:\frac{1}{2}x, \ 0:\frac{1}{2}y)$ and $(\frac{1}{2}x:x, \ \frac{1}{2}y:y)$, respectively. Then, we selected the central part of each part as the original maps. The area of the cropped maps are: $(\frac{1}{8}x:\frac{3}{8}x, \ \frac{1}{8}y:\frac{3}{8}y)$ and $(\frac{5}{8}x:\frac{7}{8}x, \ \frac{5}{8}y:\frac{7}{8}y)$.

Because a proportion maps are limited in dimensionality, the processing of image size is the main processing that filters the most original map dataset. In practice, we generally removed around 8000 map datasets within inadequate size in this step.

2. Map reformatting. A proportion of the maps accessed from massive sources are not created in a machine-readable format, and we detected and removed these unreadable maps by commonly-used data processing tools including OPENCV, Adobe Photoshop, and ArcGIS. The format we selected includes JPG, JPEG, PNG, BMP, TIFF, GIF, AI, CDR, EPS, SVG, and PSD. Moreover, for the dynamic maps generated by TIFF, we splitted this type of map into multiple frames. Then, we randomly selected one frame as the original map, since various frames in one TIFF map always shared the similar style. Finally, we converted these maps into two formats: JPG and PNG.

Since each original map are extended to two types of format, the processing of map format would double the number of maps.

3. Map color and texture filtering. Some map files might have a visually-incorrect colors and textures due to a variety of reasons, such as format conversion, data delivery, data quality, and other possible issue in terms of data quality. Since the incorrect color and texture of these maps are not suitable for visual cognition and machine intelligence techniques, we manually removed these maps. In practice, we removed 100 map datasets within inadequate size in this step.

4. Image noise filtering. Since the maps generated by unprofessional ways might contain various types of noises, it can be challenging to denoise them by automatic methods. Thus, we manually checked all maps, and removed the ones that have significant noises affecting the visualization on the map content. Specifically, we accept maps that have noise that has no obvious impact on map content since noise is inevitable in the maps accessed from massive sources. The capability of addressing noises is critical to the performance of the-state-of-the-art approaches for map pattern recognition. In practice, we only removed few map datasets within significance noises, and a majority of which are historical maps.

5. Image content filtering. It is widely acknowledged that information from the Internet and cyberspace might contain illegal or immoral clues. As the benchmark dataset would be published and shared by the global users, we manually checked all maps and removed any maps that includes potential inappropriate content. In practice, we removed around 1500 map datasets that contain illegal or immoral clues.

After map-dataset cleaning and map filtering, the total number of maps are listed in Table 2.

Table 2. Total number of maps in CartoMark

| Map category | Interpolation methods | Map format | Total map number | Total label number |
|---|---|---|---|---|
| Map Scene Classification | — | .jpg/.png | 10377 | 10377 |
| Map Text Annotation | — | .jpg/.png | 455 | |
| Map Super-resolution | Bicubic | .jpg/.png | 1100 | 1100 |
| | Gaussian | .jpg/.png | 1100 | 1100 |

**Map labelling**

The workflow of map labelling is depicted in Figure 4.. For each pre-processed map, we labelled the scene category, and generated the labelled map dataset (map scene). Then, we labelled the character proposals, and generated the labelled map dataset (text character). Finally, we respectively created two groups of low-resolution maps by bicubic interpolation and Gaussian filtering, and generated the labelled map dataset (super-resolution).

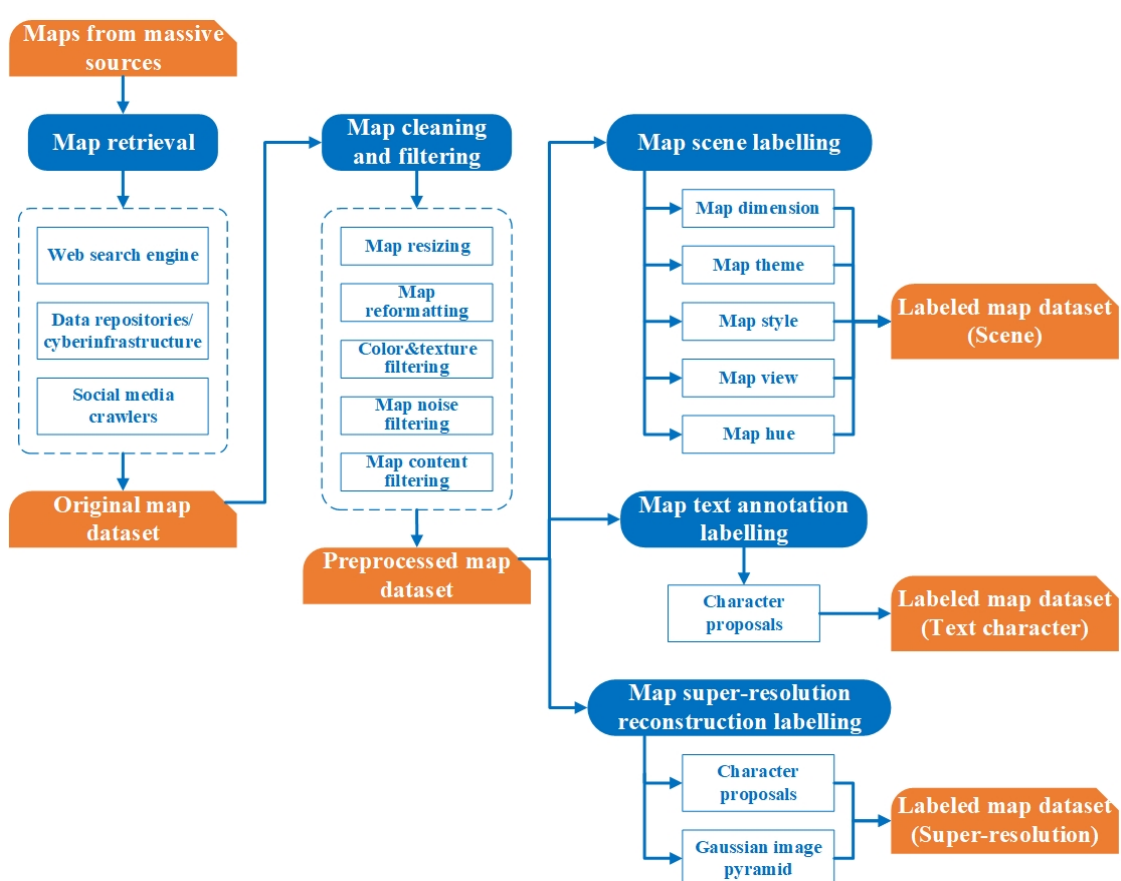

Figure 4. Workflow of map labelling.

**(1) Map scene labelling**

Based on the taxonomies defined by the cartographical classification systems and standards involving Chinese topographical mapping standard (GB/T 16820-2009: https://openstd.samr.gov.cn/bzgk/gb/newGbInfo?hcno=AADA46D2F301C30AF9103A6789C40089), USGS national maps (https://store.usgs.gov/) and land classification legend (https://www.usgs.gov/media/images/land-cover-class-legend), United Nations Maps&Geoservices (https://www.un.org/geospatial/mapsgeo/thematic), and European landscape classification (https://ec.europa.eu/eurostat/cache/metadata/en/lan_esms.htm), we design a hierarchical labelling system as shown in Figure 5a.

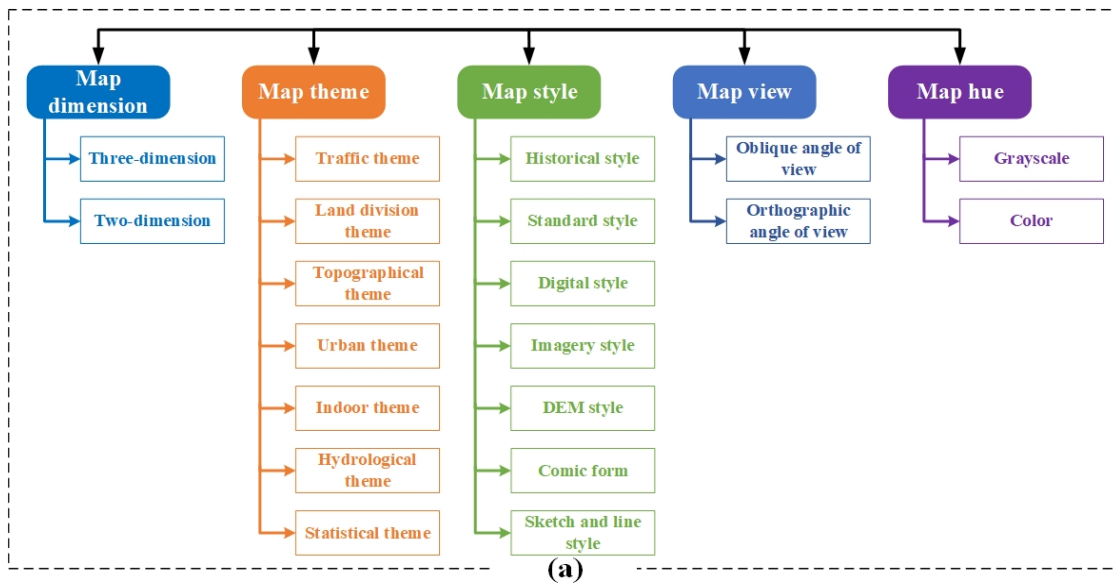

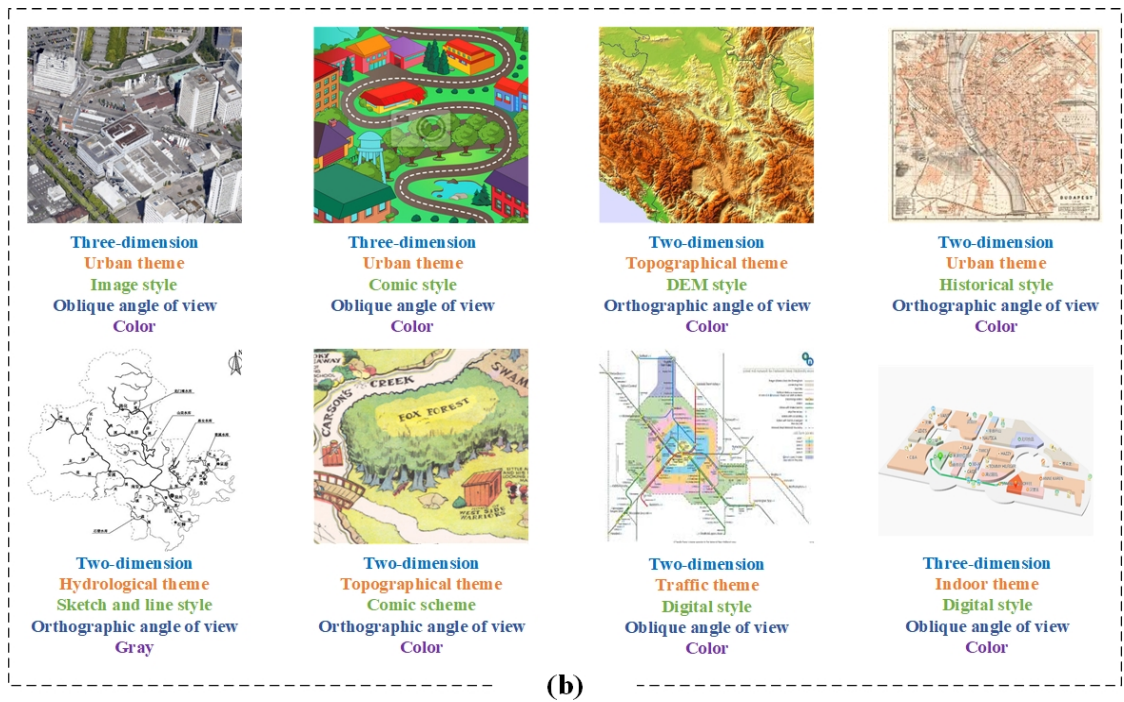

Figure 5. Hierarchical labelling system of CartoMark. a) Hierarchical labelling system, and b) the example.

The labelling system includes five hierarchies: map dimension, map theme, map scheme, map view and map hue. The details of these five hierarchies are listed in Table 3.

Table 3.

| Map dimension | Two-dimension map | A map that represents the landscape on 2D dimension. |
|---|---|---|
| | Three-dimension map | A map that represents the landscape on 3D dimension. |

| Map theme | Traffic theme | A map that represents the geospatial information regarding traffic theme. |
| --- | --- | --- |
| | Land division theme | A map that represents the geospatial information regarding separate land parcels. This type of map might be land classification map, administrative boundary map, population map, etc. |
| | Topographical theme | A map that represents the geospatial information regarding relief features and landform. |
| | Urban theme | A map that represents the geospatial information regarding urban area. This type of map might be similar to the urban theme one, but it includes other features besides traffic ones. |
| | Indoor theme | A map that represents the geospatial information regarding indoor features. This type of map might be shipping mall map, airport map, etc. |
| | Hydrological theme | A map that represents the geospatial information regarding water network and hydrological features. |
| | Statistical theme | A map that represents the gradual changes of geospatial features. This type of map might be temperature map, air quality map, quantitative pollution map, etc. |
| Map style | Historical style | A map being generated by ancient mapping ways. This type of map are always historical maps, which are printed on a paper. |
| | Standard style | A map being generated by professional mapping ways. This type of map are always topographical maps. |
| | Digital style | A map being generated by volunteered mapping ways. This type of map might be commercial maps such as Google maps, and other maps created by public. |
| | Imagery style | A map being generated by image. This type of map are always remote sensing image. |
| | DEM style | A map being generated by digital elevation models. |
| | Comic style | A map being generated by comic mapping ways. |
| | Sketch and line style | A map being generated by sketch mapping ways. This type of map always includes points and lines. |
| Map view | Oblique angle of view | A map that represents the content with oblique angle. |
| | Orthographic angle of view | A map that represents the content with orthographic angle. |
| Map hue | Grayscale | A map that represents the content by grayscale color bar. |
| | Color | A map that represents the content by color bar. |

So, each map sample would be labelled as five categories. Moreover, any individuals in each hierarchy are independent and exclusive. For example, a map could be labelled as: *{two-dimension, traffic theme, digital scheme, orthographical angle of view, color}*, but not *{two-dimension, traffic theme, topographical theme, digital scheme, orthographical angle of view,*

*color}*, as traffic theme and topographical theme belong to the same hierarchy, and are mutually exclusive. The examples are shown in Figure 5b.

**(2) Map text annotation labelling**

Map text annotation labelling includes map text annotation proposal selection and text position labelling. We selected the map text annotation proposals from the preprocessed maps by the following criteria:

a. the map text proposal has different styles (e.g. heterogeneities in character form, character glyph, character color, etc.);

b. the map text proposal has different character arrangements (e.g. curved characters, rotated characters, etc.);

c. the map text proposal has different map text backgrounds (e.g. noises, other map features overlay characters).

Then, we use a state-of-the-art labelling tool called PaddleOCR to label the positioning of each text in a map. PaddleOCR can be accessed by this link: https://github.com/PaddlePaddle/PaddleOCR. We draw a minimal bounding box for each map text, and recorded the coordinates of four corners of the bounding box. The minimal bounding box is a four-side polygon (rectangle or square), and the coordinate refers to the position of pixel that located at the corner of this minimal bounding box. Figure 6 displays the selected result of map text annotation labelling, the product of which includes the original map file and the red bounding box of each text that represents the text positions.

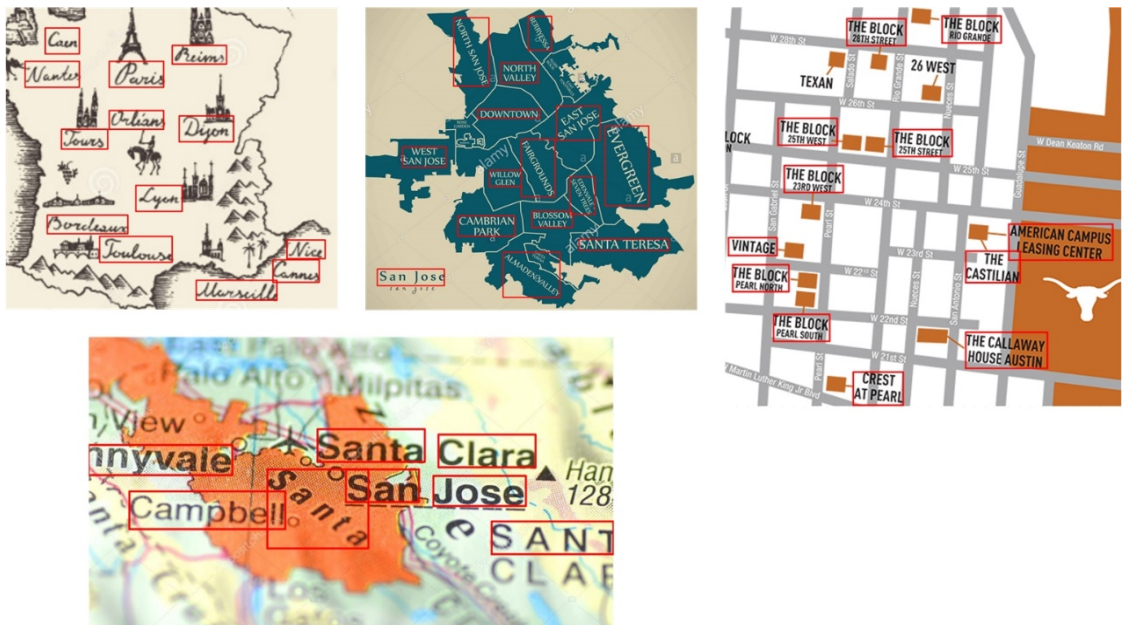

Figure 6. The example of map text annotation labelling.

**(3) Map super-resolution reconstruction labelling**

Map super-resolution reconstruction labelling includes map proposal selection and map interpolation. We selected the high-resolution maps as the map proposals from the collected maps. Then, based on each map proposal, we created its corresponding blurred maps (or low-resolution maps) by cubic spline interpolation and Gaussian filtering, respectively. Each map proposal has three low-resolution maps generated by 2X, 3X and 4X scales. The details of cubic spline interpolation can be read in Reference 23, and the function of Gaussian filtering is expressed as follows,

$$G(x, y) = \frac{1}{2\pi\sigma^2} e^{-(x^2+y^2)/2\pi\sigma^2} \quad (1)$$

where $(x, y)$ refers to the position of a pixel, $\sigma$ refers to the variance.

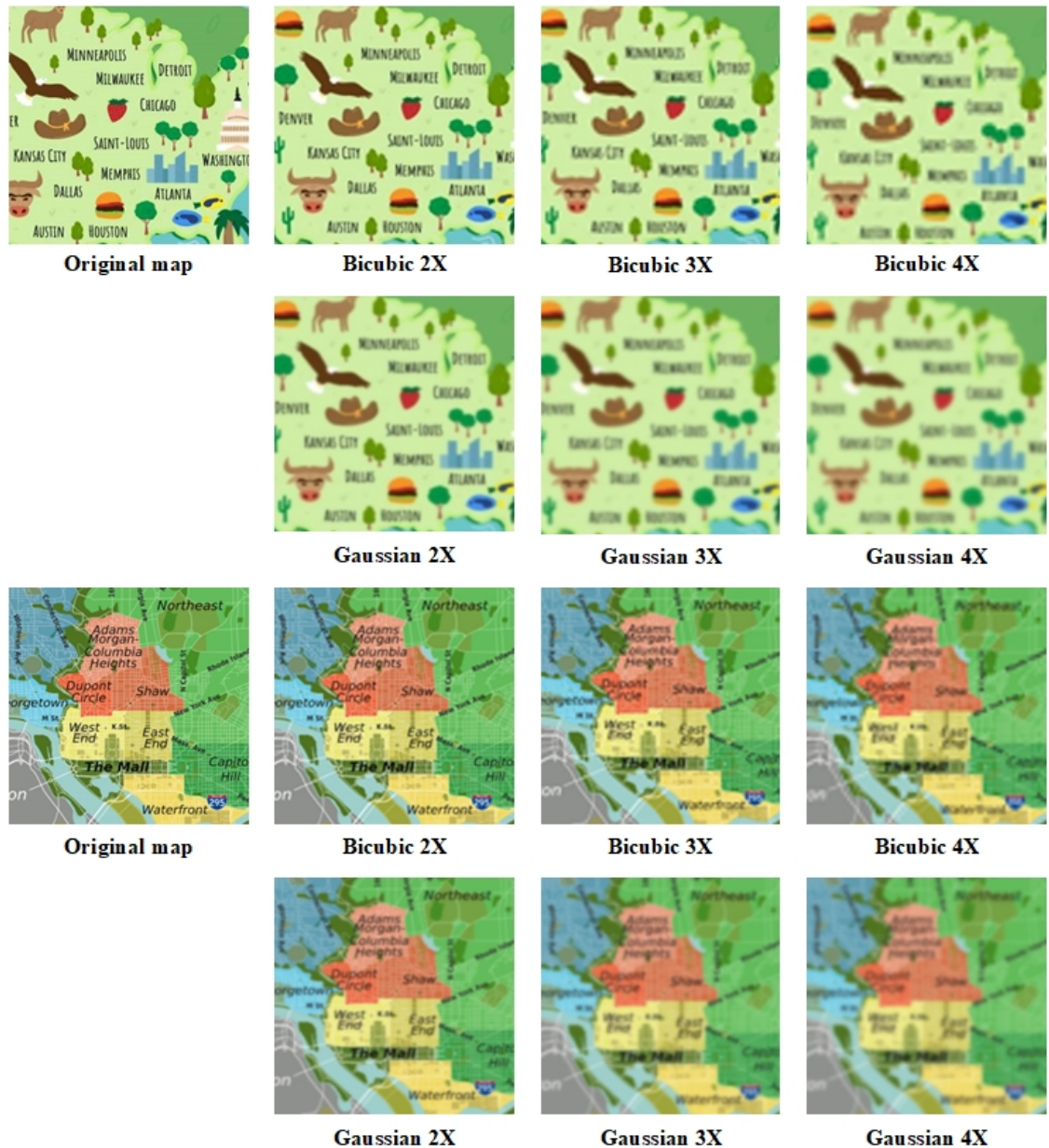

Figure 7. The example of map super-resolution reconstruction labelling.

Figure 7 illustrates the selected results of the map super-resolution reconstruction labelling process. The resulting product comprises the original map file, a set of maps containing the 2X, 3X, and 4X map files generated by Gaussian filtering and another set of files containing the 2X, 3X, and 4X map files generated by cubic spline interpolation.

**(4) Map style transferring labelling**

Map style transferring labelling includes style proposal selection and map style collection building. All collected maps for map scene classification can be used as the style proposals, and then are organized the style proposals into a map style collection. The map style collection are available for the state-of-the-art deep learning techniques regarding style transferring to change the style of maps.

## Data Records

As shown in Figure 2, the CartoMark dataset supports three tasks: map text annotation recognition, map scene classification, map super-resolution reconstruction, and map style transferring. The data modal and data format for each task are listed in Table 2.

Table 2. Data modal and data format of CartoMark

| Category | Data modal | Data format |
|---|---|---|
| Map text annotation recognition | Image | .JPG .PNG |
| | Text | .TXT |
| Map scene classification | Image | .JPG .PNG |
| | Text | .CSV |
| Map super-resolution reconstruction | Image | .JPG .PNG |
| Map style transferring | Image | .JPG .PNG |

Map text annotation recognition involves imagery modal dataset and text modal dataset. Imagery modal dataset includes map images in three commonly used formats: .JPG, and .PNG. The text modal dataset records the coordinates of each text proposal and is available in .TXT format.

Map scene classification also involves imagery modal dataset and text modal dataset. The imagery modal dataset consists of map images in the same two formats: .JPG and .PNG. The text modal dataset records the category of each map image and is available in .TXT format.

Map super-resolution reconstruction involves the imagery modal dataset alone, which records the original high-resolution map images and their corresponding low-resolution map images. The dataset is available in the two aforementioned formats: .JPG, and .PNG.

Map style transferring also involves the imagery modal dataset. The dataset includes the original map images and the maps that shares the similar content but different styles. The dataset is available in the two aforementioned formats: .JPG, and .PNG.

The benchmark dataset can be accessed by the following repositories:

• Harvard Dataverse:
https://dataverse.harvard.edu/dataset.xhtml?persistentId=doi:10.7910/DVN/ZBXJD5

• Github: https://github.com/xrzhou/CartoMark

## Technical Validation

### Original data collection validation

We used OPENCV, Adobe Photoshop, and ArcGIS to verify each collected map file, ensuring that these files are machine-readable. In the community of computer vision and pattern recognition, the benchmark dataset typically provides images in formats including .JPG and .PNG. Thus, we generated every map sample in these two formats so that machine intelligence techniques are capability of processing and addressing these map files.

Moreover, we employ the algorithms called structural similarity (SSIM) and normalized mutual Information (NMI) to measure the similarity among pre-processed maps shown in Figure 3. The SSIM is expressed as follows,

$$ssim(i, j) = \frac{(2 m_i m_j + C_1)(2 \text{cov}_{x,y} + C_2)}{(m_i^2 + m_j^2 + C_1)(s_i^2 + s_j^2 + C_2)}$$

(2)

where $i$ and $j$ are two maps, $m_i$ and $m_j$ refers to the mean of $i$ and $j$, $s_i$ and $s_j$ refers to the variance of $i$ and $j$, and $\text{cov}_{x,y}$ refers to the covariance of $i$ and $j$.

The NMI is expressed as follows,

$$nmi(i, j) = 2 \frac{\sum_i \sum_j p_{i,j} \log \frac{p_{i,j}}{p_i p_j}}{H_i + H_j}$$

(3)

where $H_i$ and $H_j$ are the information entropy of two maps $i$ and $j$, $p_i$ and $p_j$ refers to the marginal distribution of $i$ and $j$, $p_{i,j}$ refers to the joint distribution of $i$ and $j$.

For the maps identified as similar by SSIM and NMI, we then manually checked these maps and removed the similar one.

### Map labelling validation

Map labelling validation includes map scene labelling, map text annotation labelling, and map super-resolution reconstruction labelling. We designed a framework to evaluate the technical quality of each dataset by different approaches. Two university faculties, four graduate students and two undergraduate students have joined the volunteered labelling group. Since the labelling results might be varied based on different individuals, we invited these people with different education backgrounds so that diverse opinions could be considered in the labelling of each map.

#### (1) Validation of map scene labelling

Five volunteers label the category of every map. If there were at least four label results were consistent, this map would be categorized under that label. Otherwise, we would discuss about all the labelled categories and decide which category was the appropriate label. Moreover, if there was a disagreement among the volunteers, we would remove this map from the CartoMark to avoid any confusions regarding map scene. Specifically, the removed maps would be still stored, for further using in the future version of CartoMark.

### (2) Validation of map text annotation labelling

We designed a validation workflow. One volunteer labelled the text proposals of each map, and another volunteer checked the quality of the text proposal. In the first stage of checking, we examined that all text characters, which may vary in style, arrangement, and color, were accurately annotated. In the second stage of checking, we verified that the bounding box of each text proposal fully encompassed all text characters.

### (3) Validation of map super-resolution reconstruction labelling

To ensure the quality of low-resolution maps, we employed two commonly-used interpolation approaches: Gaussian image filtering and cubic spline interpolation to generate the low-resolution maps. The results generated by these two approaches have been used to evaluate the state-of-the-art machine intelligence techniques[23,24].

### (4) Validation of map style transferring labelling

To ensure the consistency of each map in the map style collection, we employed two commonly-used interpolation approaches: Gaussian image filtering and cubic spline interpolation to generate the low-resolution maps. The results generated by these two approaches have been used to evaluate the state-of-the-art machine intelligence techniques.

## Program validation

All three programs were tested by different python encoding environments including Jupiter Anaconda, Pycharm, and Pysript. Moreover, these programs were tested on different operation systems including Windows (Windows 10 and Windows 11), Mac, and Ubuntu.

## Acknowledgements
We would like to thank the valuable comments from anonymous reviewers and editors, which significantly benefit the earlier version of this manuscript.

## Code Availability
All codes of programs mentioned in this manuscript could be found in the benchmark dataset, and are available to readers without undue qualifications.

## Author contributions
X. Zhou designed the architecture of benchmark dataset, examined the labelling dataset, organized the benchmark dataset, and wrote and revised the manuscript. Y. Wen collected the original dataset and labelled the map samples, H. Li, K. Li and X. Xie labelled the map samples. Z. Yan and Z. Shao checked the architecture of benchmark dataset, and examined the labelled dataset.

## Competing interests
The authors declare no competing interests.